\def\year{2019}\relax
\documentclass[letterpaper]{article} 
\usepackage{aaai19}  
\usepackage{times}  
\usepackage{helvet}  
\usepackage{courier}  
\usepackage{url}  
\usepackage{graphicx}  
\usepackage{color}
\usepackage{subcaption}    
\usepackage{amsmath}
\usepackage{amsfonts}
\usepackage{algorithm}
\usepackage[noend]{algpseudocode}
\nocopyright
\begin{document}
%
\title{ARCHER: Aggressive Rewards to Counter bias in Hindsight Experience Replay}
\author{Sameera Lanka$^\dagger$ and Tianfu Wu$^{\dagger,\ddagger}$\\
$^\dagger$Department of ECE and $^\ddagger$Visual Narrative Initiative, North Carolina State University\\
{\tt \{slanka, tianfu\_wu\}@ncsu.edu}}

\maketitle

\begin{abstract}
Experience replay is an important technique for addressing sample-inefficiency in deep reinforcement learning (RL), but faces difficulty in learning from binary and sparse rewards due to disproportionately few successful experiences in the replay buffer. Hindsight experience replay (HER)~\cite{andrychowicz2017hindsight} was recently proposed to tackle this difficulty by manipulating unsuccessful transitions, but in doing so, HER introduces a significant bias in the replay buffer experiences and therefore achieves a suboptimal improvement in sample-efficiency. In this paper, we present an analysis on the source of bias in HER, and propose a simple and effective method to counter the bias, to most effectively harness the sample-efficiency provided by HER.  Our method, motivated by counter-factual reasoning and called ARCHER, extends HER with a trade-off to make rewards calculated for hindsight experiences numerically greater than real rewards. We validate our algorithm on two continuous control environments from DeepMind Control Suite~\cite{tassa2018deepmind} - Reacher and Finger, which simulate manipulation tasks with a robotic arm - in combination with various reward functions, task complexities and goal sampling strategies. Our experiments consistently demonstrate that countering bias using more aggressive hindsight rewards increases sample efficiency, thus establishing the greater benefit of ARCHER in RL applications with limited computing budget.     
\end{abstract}

\section{Introduction}
Humans possess the remarkable capacity to learn from experience efficiently and effectively; even from unsuccessful experience as captured by the Chinese saying -- ``failure is the mother of success". Consider how parents teach their kids to play simple block stacking games. In the early stages, parents usually provide positive feedback even when kids do not build the correct configuration. By receiving such signals, kids stay encouraged and cultivate an exploratory approach, and over time refine their skill through memory and corrective feedback. Children who learn from failure have been observed to gain faster mastery at tasks than children who are not allowed to fail~\cite{elliott1988goals}. Computationally formulating this approach will greatly help improve robot autonomy towards the ambitious goal of building machines that learn and think like people~\cite{LakeUTG16}.    

Deep reinforcement learning (RL) stands as a promising method for training robots to autonomously learn dynamic control policies~\cite{kohl2004policy,kober2011reinforcement,levine2016end}. The key advantage of deep RL lies in its capacity to learn generalized policy representations in comparison to specialized domain and task specific hand-engineered policies. However, this advantage comes with a caveat that deep RL requires a practically prohibitive amount of training data and computational time to successfully learn policies across high-dimensional, continuous state and continuous action domains, thus hindering its application in robotics. 
Deep RL methods are commonly used in conjunction with a binary success/failure reward function. Such a reward mechanism allows for the agent to discover optimal policies intended for successful completion of the task, thus eliminating the need for expert reward specification or the danger of inaccurate reward design~\cite{ng1999policy}. In high-dimensional continuous control, a characteristic property of most practical robots, a binary reward system presents a challenge when the task is complex and the goal states are rarely encountered. In these sparse reward conditions, the agent receives insufficient reinforcement to discriminate between successful and unsuccessful actions, and is consequently unable to learn an effective policy function to make progress at the task. The sparse-reward problem is particularly intensified when prolonged operation of a physical robot to explore and collect data for training is infeasible, dangerous or expensive. Thus, there exists a pressing demand to improve the sample efficiency of deep RL algorithms.

Hindsight Experience Replay (HER)~\cite{andrychowicz2017hindsight} is a method which seeks to learn from experiences that resulted in failure in addition to experiences which led to successful completion of the task. If an agent was tasked to achieve goal $A$ but ended achieving a different goal $B$, traditional RL methods consider episode as a failure and do not gain any information from this episode. However, HER proceeds as follows - in the first run, the agent stores the transitions and rewards associated with the real episode pertaining to goal $A$. Then the algorithm \textit{replays} the episode by recomputing the rewards in hindsight, under the context of achieving goal $B$ and thereby incorporates information from an unsuccessful episode to train its policy function. For off-policy RL algorithms such as DDPG~\cite{silver2014deterministic}, NAF~\cite{gu2016continuous}, DQN~\cite{mnih2015human}, incorporating HER improves sample-efficiency significantly in RL with sparse and binary rewards.  

In this paper, we challenge the central assumption underlying the vanilla HER - that the same trajectory of states and actions can be replayed in hindsight, thereby creating real and hindsight experiences which impact the learning process \textit{equivalently}, through their corresponding rewards. Our idea is inspired by the well-studied phenomenon of hindsight bias~\cite{fischhoff1975hindsight} in behavioural psychology, traditionally defined as the tendency to exaggerate the a priori predictability of outcomes after they become known. We argue that the hindsight experiences introduce bias in RL as the likelihood of replayed experience is overestimated. Our hypothesis suggests there should in fact exist a difference in the computation of rewards associated with real, generated experiences and artificially formulated hindsight experiences. We strive to quantitatively establish this difference by evaluating the relative influence of each in learning the optimal policy for a task at hand. 
 We answer the question - Does there in fact exist a hindsight bias in HER and if so, how do we mathematically justify the bias? Upon identifying the potential source of bias, we propose a solution to offset this bias using a weighted trade-off between real rewards and hindsight rewards. We empirically prove that HER works better when hindsight experiences are rewarded more (i.e., more aggressive), especially with sparse and binary rewards. We title our method \textbf{ARCHER}, \textit{Aggressive Rewards to Counter bias in HER}. In experiments, ARCHER consistently outperforms the sample-efficiency of vanilla HER as validated on two distinct continuous control domains, finger and reacher, tested in simulation using the DeepMind Control Suite~\cite{tassa2018deepmind}. 


\section{Related Work}\label{sec:relatedwork}
The high sample complexity of deep neural networks compounded with constant policy revision in RL requires that an agent must continually generate vast amount of experiences consistent with its current policy, which then ``expire" once the networks have been updated and are therefore discarded after a single use. Sample-inefficiency in deep RL remains a long-standing open challenge and many techniques have been proposed to tackle this issue.

\textit{Experience Replay} (ER), first introduced in \cite{lin1992self}, is a crucial component to stabilize convergence in off-policy deep RL networks, as demonstrated by the successful Deep Q-Network \cite{mnih2015human} algorithm. ER dissects episodes into experience quadruples of the form $(s, a, s', r)$, where the elements represent current state, action, next state and reward respectively. The experiences are stored in a collective database termed as \textit{replay buffer}, from where they undergo uniform random sampling to form minibatches for training the RL networks. In addition to allowing for temporal independence in the training set, this approach enables the reuse of past experiences, as any single experience may be sampled multiple times, and hence increases sample-efficiency. 

Subsequent research aimed to find more effective sampling strategies and compositions of the replay buffer. Prioritized Experience Replay (PER)~\cite{schaul2015prioritized} achieves higher sample efficiency by selecting experiences from the replay buffer according to a frequency distribution prioritizing the importance of each individual transition. \cite{zhang2017deeper} investigates the effect of size of the replay buffer on performance and proposes Combined Experience Replay (CER) to alleviate the liability of a large replay buffer. Hindsight Experience Replay (HER)~\cite{andrychowicz2017hindsight}, described in greater detail in the following section, strategically augments the replay buffer by reformulating unsuccessful episodes as successful transitions accomplishing a different goal. The resulting balance in successful and unsuccessful experiences in the replay buffer overcomes the disadvantage of sparse rewards.

Hindsight bias was first documented by Fischhoff \cite{fischhoff1975hindsight}, referring to the inflation in people's predicted likelihood of the true outcome of an event, after the outcome is known. This phenomenon, also termed as ``creeping determinism", is one of the most pervasive cognitive biases and routinely affects judgments in multiple domains, including medical diagnoses~\cite{arkes1988eliminating}, criminal justice~\cite{casper1989juror} and financial systems~\cite{anderson1993evaluation}.

Parallel methods to improve sample-efficiency of deep RL include learning hierarchical abstractions~\cite{kulkarni2016hierarchical,riedmiller2018learning}, reducing variance in policy gradients~\cite{gu2016q}, model-based algorithms~\cite{deisenroth2011pilco} and effective exploration~\cite{hester2013texplore,pathak2017curiosity}.

\section{ARCHER: Aggressive Rewards to Counter bias in HER}\label{sec:method}
In this section, we provide a brief mathematical introduction to Hindsight Experience Replay (HER)~\cite{andrychowicz2017hindsight} using the DDPG algorithm~\cite{silver2014deterministic} to be self-contained. Then we present our proposed method, ARCHER, in detail. 

\subsection{Background on DDPG+HER}
Let $\mathcal{S}$ and $\mathcal{A}$ denote the state-space and the action-space of the environment respectively, and let $\mathcal{G}$ represent the space of goals we wish to achieve. We assume that for every state $s \in \mathcal{S}$, there exists some goal $g \in \mathcal{G}$ achieved in that state. This gives rise to the mapping $m: \mathcal{S} \rightarrow \mathcal{G}$ where $m(s) = g$ denotes the achieved goal in state $s$. Moreover, we state that every goal $g\in \mathcal{G}$ corresponds to the predicate $f_g: \mathcal{S}\rightarrow \{0,1\}$, which indicates whether the goal $g$ has been realized in state $s$. The true objective of an agent is to reach a state $s$ such that $f_g(s)=1$, following which it receives a successful reward signal from the environment. 

In environments with complex tasks and sparse rewards, it is extremely unlikely that the agent achieves the goal. Hence, for most of the encountered states, $f_g(s)\neq 1$ and the agent largely only receives unsuccessful rewards. To solve the deficit of successful experiences, and encourage the agent to effectively discriminate between good and bad policies, we exploit the knowledge that although the agent has not achieved the pre-specified goal $g$, it has achieved goal $g^h = m(s)$, and by extension, $f_{g^h}(s) = 1$. We call $g^h$ the hindsight goal and we now explain how to incorporate HER into the off-policy RL algorithm, Deep Deterministic Policy Gradient (DDPG).

DDPG architecture consists of two neural networks - an actor $\mu(s; \theta^\mu)$, representing a deterministic policy function parameterized by $\theta^\mu$ and a critic $Q(s, a; \theta^Q)$ which calculates the Q-value of state-action pairs and is parameterized by $\theta^Q$; for all $a\in \mathcal{A}$ and $s\in \mathcal{S}$. 

To train the actor and critic networks using HER, the state inputs to the networks are appended with the desired goal for the episode. Let $s||g$ denote the state vector concatenated with the goal. We write the actor and the critic equations as,
\begin{eqnarray}
  a_t = \mu(s_t||g; \theta^\mu), \\
  Q_t = Q(s_t||g\;, a_t;  \theta^Q)
\end{eqnarray}
To stabilize training, we maintain two gradually updated duplicates of the original actor and critic networks, denoted as target actor $\mu'(a; \theta^{\mu'})$ and target critic $Q'(s, a; \theta^{Q'})$. The target networks are not optimized directly by gradient descent and are only used to compute the target Q-values Eqn.~\eqref{eq:targetvalue} for the critic loss Eqn.~\eqref{eq:loss_critic}. 

Training DDPG using HER consists of two phases. In the first phase, we sample a target goal $g$ and an initial state $s_0$ at the start of every episode. Then for every time step $t=0, \cdots, T-1$, where $T$ is the length of the episode, we run the policy network and add the generated real experience tuples, $e_t$, to the replay buffer.
\begin{equation}
e_t = (s_t||g\,,a_t\,,s_{t+1}||g\,, r_t), \label{eq:real}
\end{equation}
where $a_t=\mu(s_t||g; \theta^{\mu}) + \mathcal{N}$, i.e. adding noise to the actor output to facilitate exploration, and $r_t$ reward is computed by a reward function dependent on the state transitions and goal. For example, if we adopt negative reward for failure and we have, 
\begin{equation}
r_t = r(s_t, a_t, g) = - [f_{g}(s_{t+1})=0]
\end{equation}
This phase is known as standard experience replay. In the second phase, we execute hindsight experience replay where we pretend that the goal we intended to achieve was the goal corresponding to the state of the environment at the terminal step of the episode $g^h = m(s_T)$ (other strategies can be used, see~\cite{andrychowicz2017hindsight}). We modify the sequence of transitions generated during standard experience replay, by replacing the real goal $g$ with the achieved goal $g^h$. We supplement the replay buffer with hindsight experience tuples,
\begin{equation}
e^h_t = (s_t||g^h\,,a_t\,,s_{t+1}||g^h\,, r^h_t), \label{eq:fake}
\end{equation}
where the reward $r^h_t= - [f_{g^h}(s_{t+1})=0]$ is recomputed at each step in accord with the fake goal $g^h$. 

Thus, the replay buffer is augmented with additional training data containing successful reinforcement signals, mitigating the sparse reward problem. The successful transitions which accompany the HER increase the sample-efficiency of DDPG and help the agent learn policies for high-dimensional continuous control.

The actor and critic are optimized by stochastic gradient descent~\cite{rumelhart1986learning} by uniformly sampling mini-batches from the replay buffer. Given a mini-batch consisting of $n$ experience tuples $\{e_i=(s_i||g, a_i, s_{i+1}, r_i)\}_{i=1}^n$ (each $e_i$ can be either a real experience or hindsight experience), the critic network $Q(s||g, a; \theta^Q)$ is updated by minimizing the critic loss between the predicted Q-value and the temporal difference target~\cite{sutton1988learning}, 
\begin{equation}
L_{critic} = \frac{1}{n} \sum_{i=1}^n [y_i - Q(s_i||g, a_i; \theta^Q)]^2, \label{eq:loss_critic}  
\end{equation}
where the target value $y_i$ is computed by the target networks, 
\begin{equation}
y_i = r_i + \gamma Q'(s_{i+1}||g, \mu'(s_{i+1}||g;\theta^{\mu'}); \theta^{Q'}), \label{eq:targetvalue}
\end{equation} 
and $\gamma$ is a predefined discount factor. The actor $\mu(s||g; \theta^{\mu})$ is trained to maximize the output of the critic by minimizing the following loss using deterministic policy gradient~\cite{silver2014deterministic},
\begin{equation}
L_{actor} = -\frac{1}{n} \sum_{i=1}^n Q(s_i||g, \mu(s_i||g;\theta^{\mu}); \theta^Q). 
\end{equation}
The target networks are updated in a conservative way by,
\begin{eqnarray}
\theta^{Q'}\leftarrow \tau \theta^Q + (1-\tau)\theta^{Q'}, \\
\theta^{\mu'}\leftarrow \tau \theta^{\mu} + (1-\tau)\theta^{\mu'}, 
\end{eqnarray}
where $\tau$ is a small positive real value.

\begin{algorithm}[t]
	\caption{ARCHER}
    \label{alg:wrHER}
	\begin{algorithmic}
	\Statex \textbf{Given}:
    \Statex \hspace{4ex} $\bullet$ an off-policy RL algorithm $\mathbb{A}$,\Comment{e.g. DDPG}
    \Statex \hspace{4ex} $\bullet$ a strategy $\mathbb{S}$ for sampling goals for replay
    \Statex \hspace{4ex} $\bullet$ a reward function $r : \mathcal{S}\times\mathcal{A}\times\mathcal{G}\rightarrow\mathbb{R}$
    \Statex \hspace{4ex} $\bullet$ real reward weight $\lambda_r$, hindsight reward weight $\lambda_h$
    \Statex Initialize $\mathbb{A}$
    \Statex Initialize replay buffer $R$
    \For{episode = 1, $M$}
    	\State Sample a goal $g$ and initial state $s_0$
        \For{$t = 0, T-1$}
        	\State Sample an action $a_t$ using the behavior policy from $\mathbb{A}$:
            \State \hspace{4ex} $a_t \leftarrow \mu(s_t||g\,; \theta^\mu)+\mathcal{N}$ 
            \State Execute the action $a_t$ and observe a new state $s_{t+1}$
        \EndFor
    	\For{$t = 0, T-1$}
        	\State $r_t = \lambda_r\times r(s_t, a_t, g)$
        	\State Store the transition $(s_t||g, a_t, r_t, s_{t+1}||g)$ in $R$ 
            \State Sample a set of additional goals for replay, $\mathcal{G} = \mathbb{S}(\mbox{current episode})$
            \For{$g^h \in \mathcal{G}$}
            	\State $r^h_t = \lambda_h\times r(s_t, a_t, g^h)$ 
                \State {Store the transition $(s_t||g^h, a_t, r^h_t, s_{t+1}||g^h)$ in $R$} \Comment{HER with weighted rewards}
            \EndFor
        \EndFor
        \For{$t = 1, N$}
        	\State Sample a minibatch $B$ from the replay buffer $R$
            \State Perform one step of optimization using $\mathbb{A}$ and the minibatch $B$
        \EndFor
    \EndFor
	\end{algorithmic}
\end{algorithm}

\subsection{The Proposed Method: ARCHER}
In this section, we first examine the source of bias in HER, and then present our algorithm ARCHER which uses more aggressive rewards for hindsight experiences to combat the bias, and thus achieving greater sample-efficiency. 

Compare the real experience tuple $(s_t||g, a_t, s_{t+1}||g, r_t)$ in \eqref{eq:real} to the artificially constructed hindsight experience tuple $(s_t||g^h, a_t, s_{t+1}||g^h, r^h_t)$ in \eqref{eq:fake}. This conversion of the a real experience to its corresponding hindsight experience makes the following unjustified assumption - \emph{Given different inputs $s_t||g$ and $s_t||g^h$, the policy network $\mu(\cdot; \theta^\mu)$, returns the \text{same} action, $a_t$}. This assumption overestimates the probability assigned by the policy network to $a_t$, given the input $s_t||g^h$. If we actually execute the policy network with $s_t||g^h$ as input, it is unlikely to output $a_t$, making $s_{t+1}$ also unlikely. Hence we observe a chain of compounding uncertainty along the hinsight episode. 

Therefore, to more effectively use HER, we require to correct the hindsight bias induced by this overestimated probability. The intuitive check would be to generate hindsight experiences by using models capable of counterfactual reasoning, i.e. by asking the network \textit{what if $g^h$ was the actual goal}, instead of mere substitution of real experiences. However, this a critical limitation of deductive learning models~\cite{pearl2018theoretical} and remains a challenge for the future.

We propose a simple solution to offset this bias. We make that case that a hindsight experience and a real experience cannot be treated in the same manner as real experiences are authentically generated by interacting with the environment, and hence their probability is unbiased. In contrast, to overcome hindsight bias, we need to match the true probability of the hindsight experiences to their biased probability. To do so, we nudge the current policy to be more consistent with the hindsight data in the replay buffer. Hence, to meet the overestimated hindsight likelihood of $a_t$ for $s_t||g^h$, we utilize more aggressive hindsight rewards, so that a large positive reward given to a successful hindsight transition greatly increases the Q-value of the hindsight state-action pair, which indirectly drives an aggressive policy update towards choosing this maximizing action for the given hindsight state.    

We test our hypothesis by introducing two real-valued scalar multipliers, $\lambda_r$ and $\lambda_h$, to distinguish between real rewards $r_t$ and hindsight rewards $r^h_t$ as follows:
\begin{eqnarray}
  r_t = \lambda_r \times r(s_t, a_t, g)\label{eq:realrew}\\ 
  r^h_t = \lambda_h \times r(s_t, a_t, g^h) \label{eq:hsrew}
\end{eqnarray}
where r($\cdot$) is the given reward function for the task.

We refer to $\lambda_r$ and $\lambda_h$ as trade-off parameters as they proportionally increase the value of one category of reward with respect to the other. We investigate the impact of this weighted reward mechanism on finding the optimal policy. Vanilla HER is a special case with $\lambda_r=\lambda_h=1$. 

\begin{figure*} [!ht]
	\begin{subfigure}{1\textwidth}
    	\centering
		\includegraphics[width=.9\textwidth]{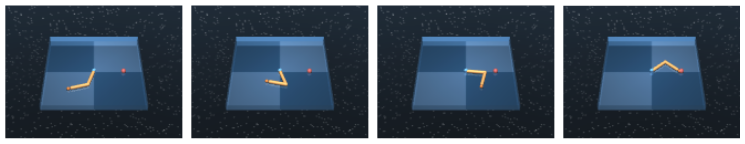}
		\label{fig:reacher}
	\end{subfigure} \\
	\begin{subfigure}{1\textwidth}
    	\centering
		\includegraphics[width=.9\textwidth]{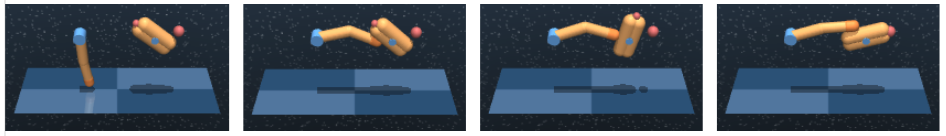}
		\label{fig:finger}
	\end{subfigure}
\caption{Illustration of the two environments: Reacher (Top) and Finger (Bottom). See text for details. }
\label{fig:env} \vspace{-2mm}
\end{figure*}

ARCHER framework requires that $r^h_t \geq r_t$. Hence using Eqn.~\eqref{eq:realrew} and Eqn.~\eqref{eq:hsrew} we get,
\begin{eqnarray}
  \lambda_h \times r(s_t, a_t, g^h) > \lambda_r \times r(s_t, a_t, g)
\end{eqnarray}

\textbf{Therefore in domains with positive reward functions, i.e. $r(\cdot) \geq 0$, ARCHER comprises trade-offs with $\lambda^r < \lambda^h$. Conversely, in negative reward functions, i.e. $r(\cdot) \leq 0$, ARCHER comprises with trade-offs $\lambda^r > \lambda^h$.}

Using trade-off, the target values for real experience and hindsight experience (Eqn.~\ref{eq:targetvalue}) can be rewritten as, 
\begin{eqnarray}
y_i = \lambda_r\cdot r_i + \gamma Q'(s_{i+1}||g, \mu'(s_{i+1}||g;\theta^{\mu'}); \theta^{Q'}),\\
y_i^h = \lambda_h\cdot r_i^h + \gamma Q'(s_{i+1}||g^h, \mu'(s_{i+1}||g^h;\theta^{\mu'}); \theta^{Q'})
\end{eqnarray}

\section{Experiments}\label{sec:exp}
In this section, we present our experimental analyses and ablation studies, along with the corresponding inferences.

\subsection{Simulation Environments}
We evaluate our method on the DeepMind (DM) Control Suite~\cite{tassa2018deepmind} simulation software. This library consists of a set of continuous control environments in Python, built on top of the MuJoCo physics engine~\cite{todorov2012mujoco}. Each environment in the suite provides a physics task along with a well-defined continuous action space $\mathcal{A}$, continuous state/observation space $\mathcal{S}$, and intrinsic transition dynamics based on the physics engine. For our experiments, we program our own reward functions to conduct ablation studies on ARCHER and verify its robustness, as detailed in the following sections.
We tested our algorithm on the following two domains:

\subsubsection{Reacher}
The Reacher environment (Fig. \ref{fig:env} (a)) includes a 2-DoF planar robot arm where the agent must place its end effector at a randomized target, indicated by the red sphere. The state input $s_t$ is a 4-element vector with the first 2 elements containing the generalized positions and next 2 elements containing the generalized velocities of the two arm joints, as encoded by the Mujoco physics engine. $s_t$ is concatenated with a real/hindsight goal $g$ which specifies the 3-dimensional global coordinate of the target sphere/end effector respectively. The action $a_t$ is a 2-dimensional vector where each element informs the relative displacement of each joint from its current position.

\subsubsection{Finger}
The finger environment (Fig. \ref{fig:env} (b)) is a multi-body arrangement where a planar 2-DoF robot arm has to flick a spinner resting on an unactuated hinge so as to place the red tip of the spinner on the target indicated by a red sphere. The arm must therefore learn a policy in a environment with discontinuous dynamics~\cite{tassa2010stochastic}. The state $s_t$ is concatenation of a 4-dimensional position vector (generalized joint points as the first two elements and the relative (x, z) position of the spinner tip to its hinge as the next two elements), a 3-dimensional velocity vector of the 3 joints and followed by signals from the two touch-sensors on the top and bottom of the spinner. This 9-dimensional state is concatenated with a goal $g$ determined by the position of target sphere relative to the hinge for real episodes, and the relative position of the spinner tip for hindsight episodes. The 2-dimensional action vector $a_t$ specifies the relative displacement of the two arm joint from their current positions.

In both domains, the initial state and target locations are randomly initialized at the beginning of every episode.   
\subsection{Reward Functions}
\label{sec:reward}
We analyze the performance of ARCHER in comparison to vanilla HER, across 3 types of rewards for each task:
\begin{enumerate}
  \item \textbf{Binary -1/0 reward}: In this case, the agent receives a reward of -1 for every time-step in the episode where the goal is not achieved, and receives 0 when the agent it successful. 
    \begin{eqnarray}
      \nonumber r(s_t, a_t, g)& = & -[f_g(s_{t+1})=0] \\ 
      &=&\begin{cases}
        -1, &\mbox{ if } m(s_{t+1}) \neq g\\
        0, &\mbox{ otherwise }
    \end{cases} 
    \label{eq:neg_reward}
\end{eqnarray}
This reward function penalizes the agent for not achieving the goal at every time-step and therefore the agent is incentivized to learn a time-efficient optimal policy. The negative reward punishes the agent for executing unproductive actions. It is used in the vanilla HER.

\item \textbf{Binary 0/+1 reward}: In this case, the agent is awarded a value of 0 for every unsuccessful time-step and is granted a reward of 1 when the goal is achieved.
\begin{eqnarray}
  \nonumber r(s_t, a_t, g) & = & [f_g(s_{t+1})=1] \\ 
    &=& \begin{cases}
    0, \mbox{ if } m(s_{t+1}) \neq g\\
    1, \mbox{ otherwise }
    \end{cases} 
    \label{eq:pos_reward}
\end{eqnarray}
The positive rewards encourage the agent to learn the policy which allows it to collect most reward and encourages the successful actions taken by the agent. 

\item \textbf{Shaped reward}: In this case, the agent is provided with a continuous real-valued reward signal, in proportion to some metric representing how close to/far away from success the agent is. For our experiments, we have selected the negative of Frobenius norm of the difference vector between the achieved goal and actual goal. The Frobenius norm of a vector $A$ is given as \[\| A \|_F = \left( \sum_{i,j=1}^n | a_{ij} |^2 \right)^{1/2}.\] 
Therefore, the shaped reward function is given by 
\begin{eqnarray}
  r(s_t, a_t, g) = - \|m(s_{t+1}) - g \|_F \label{eq:shaped}
\end{eqnarray}

The agent receives a large negative reward for states far away from the goal state and rewards closer to 0 when the agent is close to the goal. Vanilla HER performs poorly in tasks with shaped reward.
\end{enumerate}

\begin{figure*} [!htbh]
	\begin{subfigure}{1\textwidth}
    	\centering
		\includegraphics[width=.9\textwidth]{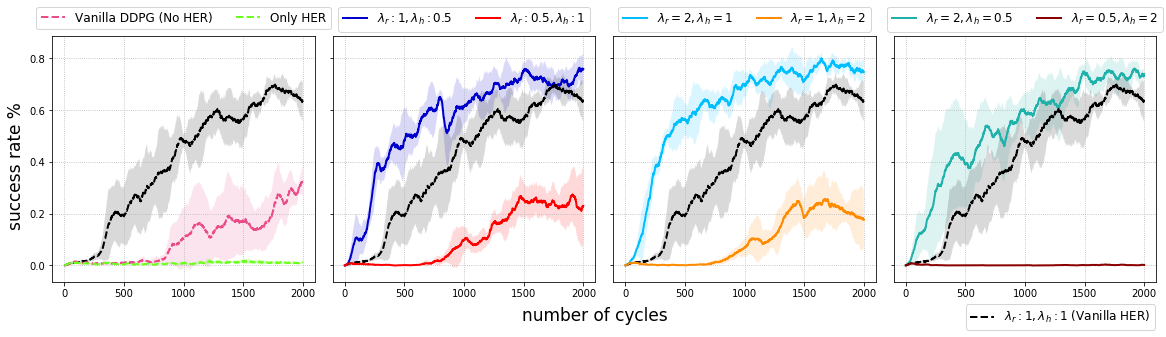}
	\end{subfigure} \\ 
	\begin{subfigure}{1\textwidth}
    	\centering
		\includegraphics[width=.9\textwidth]{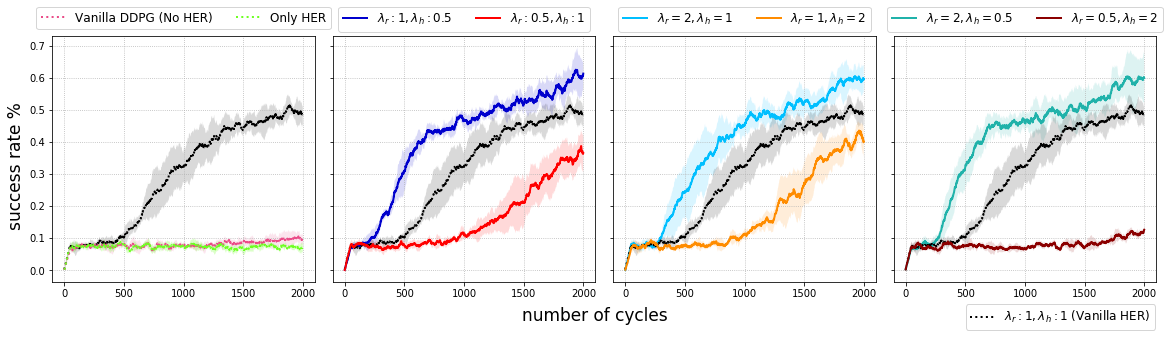}
	\end{subfigure}
\caption{Policy performance in the Reacher (Top) and the Finger (Bottom) environment with sparse binary negative rewards and the \textit{final} sampling strategy for hindsight goals. Best viewed in color.}
\label{fig:neg_final_result} \vspace{-2mm}
\end{figure*}

\begin{figure*} [t]
	\begin{subfigure}{1\textwidth}
    	\centering
		\includegraphics[width=1\textwidth]{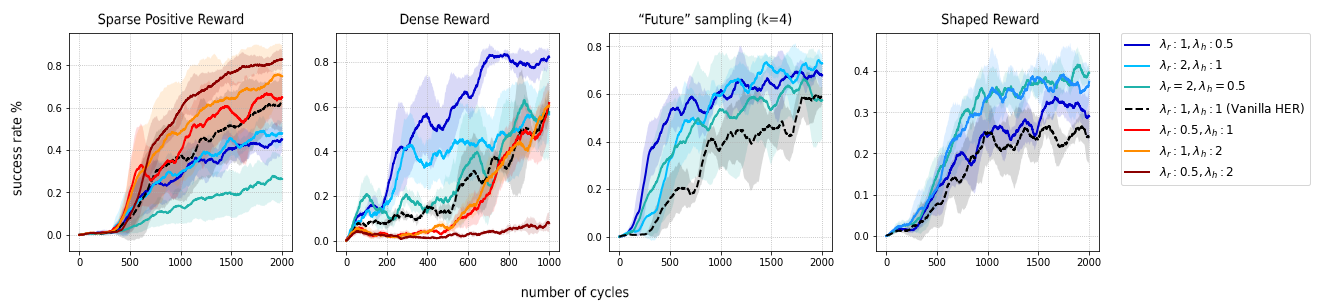}
	\end{subfigure} \\
	\begin{subfigure}{1\textwidth}
    	\centering
		\includegraphics[width=1\textwidth]{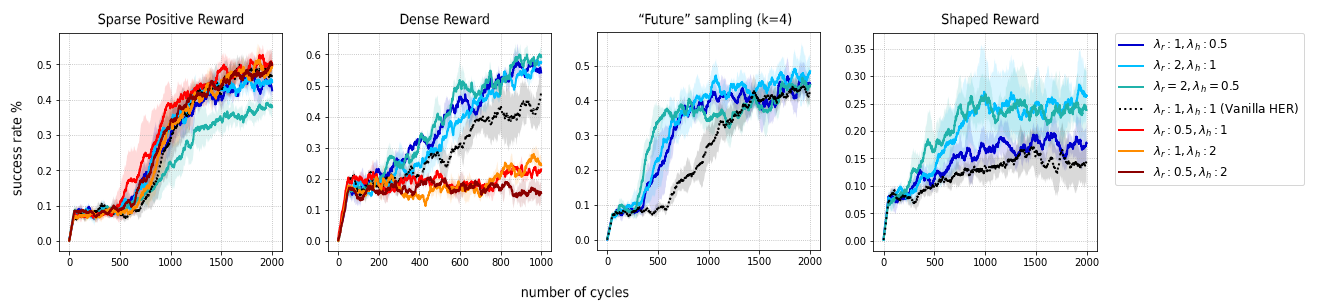}
	\end{subfigure}
\caption{Ablation studies in the Reacher (Top) and the Finger (Bottom) environment. See text for detail. Best viewed in color.}
\label{fig:ablation} \vspace{-2mm}
\end{figure*}

\textbf{Strategy for sampling goals for experience replay:} We test two strategies  proposed in HER~\cite{andrychowicz2017hindsight}: first is the \textit{final} strategy which replays with the hindsight goal corresponding to the final state in each episode, and the other is the \textit{future} strategy which replays with $k$ random states which come from the same episode as the transition being replayed and were observed after it (we use the best reported practice with $k=4$).

\subsection{Trade-off Parameters}
For each of our experiments, we carefully selected the trade-off parameters to gain insight into the relative reward optimization between hindsight and standard experience replay. The following set of weights helps us understand the impact on performance driven by (i) the ratio between hindsight and real reward weights (ii) the magnitude of these weights relative to baseline HER.

\begin{enumerate}
\item $\lambda_r = 1, \lambda_h = 1$: These weights are the same as vanilla-HER, which serves as the baseline standard for our tests.

\item $\lambda_r, \lambda_h \in \left\{0.5, 1\right\}$: The magnitude of the smaller weight is half of the baseline weight. We have $\lambda_r = 0.5, \lambda_h = 1$ and $\lambda_r = 1, \lambda_h = 0.5$. 

\item $\lambda_r, \lambda_h \in \left\{1, 2\right\}$: The magnitude of the larger weight is twice the baseline weight. We have $\lambda_r = 2, \lambda_h = 1$ and $\lambda_r = 1, \lambda_h = 2$. 

\item $\lambda_r, \lambda_h \in \left\{0.5, 2\right\}$: The magnitude of the larger weight is twice the baseline weight and the magnitude of the smaller weight is half of the baseline weight. We have $\lambda_r = 2, \lambda_h = 0.5$ and $\lambda_r = 0.5, \lambda_h = 2$. 
\end{enumerate}

\subsection{Network Architecture and Training}
For our experimental setup, the actor and critic networks were designed with 2 fully connected hidden layers, consisting of 400 ReLU~\cite{nair2010rectified} neurons in the first and 300 ReLU neurons in the second layer. The output layer of the actor networks used $\tanh$ activation. The layers of the networks were initialized uniformly from $\left[\frac{-1}{\sqrt f}, \frac{1}{\sqrt f}\right]$ where $f$ was the fan-in of the layer. The final layers of the networks were initialized uniformly in the range $[-3\times10^{-3},\,3\times10^3]$. The discount factor $\gamma$ was 0.98. The weight $\tau$ for the weighted update of the target networks was 0.001. To encourage exploration, we added Ornstein-Uhlenbeck process~\cite{uhlenbeck1930theory} noise with $\theta = 0.15$ and $\sigma = 0.2$ to the output of the actor. The added noise is multiplied by a factor $\epsilon$, with an initial value of 0.1 with an exponential decay factor of 0.99, to gradually reduce exploration. The learning rates for the actor and critic were $10^{-4}$ and $10^{-3}$ respectively, and the networks were trained using Adam~\cite{kingma2014adam} optimization. We used a replay buffer of size $10^5$ from which minibatches of size 128 were uniformly sampled for training. The networks were trained for 2000 cycles, where 1 cycle represents running the policy for 16 episodes followed by 40 steps of optimization.

\subsection{Results}
We evaluated our method by comparing the success rate of the learned policies against the required number of cycles to achieve that performance. An episode was considered successful if on final step of the episode, the end effector was placed at the target in the case of Reacher; or the spinner tip was aligned with the target in the case of Spinner. Each episode consisted of 50 MuJoCo time-steps. The results presented in Fig.~\ref{fig:neg_final_result} and Fig.~\ref{fig:ablation} show the performance curves of the actual actor networks of each agent, averaged over 5 random seeds and smoothed across the past 50 elements. The blue tinted plots depict HER with $\lambda^r > \lambda^h$ tradeoff while the orange/red tinted plots depict HER with $\lambda^r < \lambda^h$ tradeoff. Vanilla HER is shown with a perforated black plot.

\subsubsection{Experiment I: Using Sparse Negative Rewards and the \textit{Final} Goal Sampling Strategy.} 

Fig.~\ref{fig:neg_final_result} displays the final performance plots in the Reacher and Finger domains with a binary negative -1/0 reward function \eqref{eq:neg_reward}. In both domains, we observe from the leftmost graph that vanilla HER greatly improves the performance of DDPG. We also see that learning exclusively from only real or only hindsight experiences decreases performance. Hence we need a combination of real and hindsight experiences for competent learning. The following 3 graphs show that the fastest performance is demonstrated in the curves where $\lambda^r > \lambda^h$, representing ARCHER. Moreoever, when we specify trade-offs to exacerbate the discrepancy between the true hindsight probability and biased probability ($\lambda^r < \lambda^h$), performance is adversely affected and sample-efficiency decreases. This confirms our hypothesis of hindsight bias in HER.

\subsubsection{Experiment II: Ablation Studies}

Fig.~\ref{fig:ablation} shows the results of our ablation studies. 
\begin{enumerate}
\item The first column shows the performance curves for the different algorithms when presented with a sparse binary positive 0/+1 reward function \eqref{eq:pos_reward}. The striking observation is that the orange tinted curves are above the baseline while the blue curves fall below. This result is consistent with ARCHER as when the reward function is positive, $\lambda^r < \lambda^h$ ensures that hindsight rewards are numerically greater. This graph shows that ARCHER is not just a coincidence of implicit hyper-parameter tuning such as learning rate increase, but is robust to changes in reward sign.

\item In the second column we illustrate the effectiveness of ARCHER in dense binary negative reward condition, where the size of the target sphere in both domains is magnified. As shown in the graph, the blue plots depicting ARCHER perform better than vanilla HER. In Reacher, the other curves eventually catch up to ARCHER but in the high-dimensional Finger domain, ARCHER maintains a noticeable lead. Hence the main benefit of ARCHER lies in high-dimensional, sparse reward domains.

\item In the third column, we plot the results for vanilla HER and HER with trade-offs when 4 "future" goals are sampled for hindsight replay with negative binary sparse reward. In the fourth column, we show the performance of the different algorithms when provided negative shaped reward~\eqref{eq:shaped}. Under both arrangements, we observe that the blue ARCHER curves deliver the most sample efficiency. Shaped reward was mentioned as a weakness for vanilla HER in ~\cite{andrychowicz2017hindsight}. Here, we see that ARCHER does outperform the baseline even with shaped reward, although the final performance is lower than the other reward functions.
\end{enumerate}

\section{Conclusion and Discussion} \label{sec:conclusion}
In this paper, we identify bias in hindsight experience replay and present a method ARCHER to counter the bias and increase sample-efficiency of deep RL algorithms, using numerically greater hindsight rewards. We also empirically verified our hypothesis of hindsight bias in vanilla HER by exhibiting that when DDPG is trained with rewards opposite to those specified by ARCHER, the bias is amplified and performance degrades. Using the Finger and Reacher simulation environments from DeepMind control suite, we demonstrated the increased sample efficiency derived from ARCHER in comparison to baseline vanilla HER. Our ablation studies prove ARCHER consistently outperforms vanilla HER across various reward functions and task complexities. Hence, ARCHER grants reliable sample-efficiency especially in continuous control and robotic with limited computing budget.

A few interesting directions emerge for further exploration. Some of our experiments reveal that ARCHER enjoys higher sample-efficiency only until a context-dependent number of samples, after which vanilla HER catches up to ARCHER. This effect makes intuitive sense as the high performance of ARCHER leads to the fast convergence of real and hindsight experiences, and diminished hindsight bias. Hence, a scheduled annealing of ARCHER remains of interest.
Also, we specifically constructed a simple linear relation to derive a more informative hindsight reward function, however we believe that there may exists a more complex mapping between real and hindsight rewards and hence it may be advantageous to introduce a generative model to learn the latent mapping.
Moving beyond predefined trade-off parameters and limited number of their combinations will allow us to construct a more mathematically rigorous theory of hindsight bias. We did not deploy ARCHER to real robots, and will investigate this in the future. 

\bibliographystyle{aaai}
\bibliography{references}
\end{document}